\documentclass{article}
\usepackage{amssymb}
\usepackage{xcolor}
\usepackage[implicit=false]{hyperref}

\long\def\comment#1{}

\newtheorem{assumption}{Assumption}
\newtheorem{theorem}{Theorem}

\newcommand{\Thm}[1]{Theorem~\ref{#1}}

\newcommand{\Ass}[1]{Assumption~\ref{#1}}

\newcommand{\Eq}[1]{Equation~\ref{#1}}
\newcommand{\Eqs}[1]{Equations~\ref{#1}}
\newcommand{\qed}{\mbox{$\square$}}
\newcommand{\Sec}[1]{Section~\ref{#1}}

\newcommand{\X}{X}
\newcommand{\x}{x}
\newcommand{\Db}{D}
\newcommand{\U}{{\bf X}}
\newcommand{\C}{{\bf x}}
\newcommand{\bY}{{\bf Y}}
\newcommand{\by}{{\bf y}}
\newcommand{\Pai}{{\bf Pa}_i}
\newcommand{\pai}{{\bf pa}_i}
\newcommand{\pail}{{\bf pa}_{il}}
\newcommand{\PaSi}{{\bf Pa}^S_i}
\newcommand{\paSi}{{\bf pa}^S_i}
\newcommand{\PaScii}{{\bf Pa}^{S_{ci}}_i}

\newcommand{\Bs}{S}
\newcommand{\Bsc}{S_c}
\newcommand{\Bsci}{S_{ci}}

\newcommand{\Bsone}{S_1}
\newcommand{\Bstwo}{S_2}

\newcommand{\Bxtoy}{S_{x \rightarrow y}}

\newcommand{\Bxy}{S_{xy}}

\newcommand{\hBs}{S^h}
\newcommand{\hBsc}{S^h_c}
\newcommand{\hBsci}{S^h_{ci}}

\newcommand{\hBsone}{S^h_1}
\newcommand{\hBstwo}{S^h_2}

\newcommand{\hBxtoy}{S^h_{x \rightarrow y}}
\newcommand{\hBytox}{S^h_{x \leftarrow y}}
\newcommand{\hBxy}{S^h_{xy}}

\newcommand{\dXdY}[2]{\partial {#1} / \partial {#2}}

\newcommand{\xtoy}{X \rightarrow Y}

\newcommand{\Th}{\mbox{\boldmath $\theta$}}
\newcommand{\ThU}{\Th_{\bf x}}
\newcommand{\Thi}{\Th_i}
\newcommand{\TBs}{\Th_{s}}
\newcommand{\TBsc}{\Th_{sc}}
\newcommand{\TBsci}{\Th_{sci}}
\newcommand{\TBsone}{\Th_{s1}}
\newcommand{\TBstwo}{\Th_{s2}}

\newcommand{\TBxtoy}{\Th_{x \rightarrow y}}
\newcommand{\TBytox}{\Th_{x \leftarrow y}}
\newcommand{\TBxy}{\Th_{xy}}
\newcommand{\ta}[1]{\theta_{#1}}
\newcommand{\tijk}{\theta_{ijk}}

\newcommand{\CTh}{\Theta}
\newcommand{\CTBs}{\Theta_{s}}
\newcommand{\CTBsone}{\Theta_{s1}}
\newcommand{\CTBstwo}{\Theta_{s2}}

\newcommand{\vecm}{{\bf m}}
\newcommand{\vecv}{{\bf v}} 
\newcommand{\vecb}{{\bf b}} 
\newcommand{\matT}{T} 
\newcommand{\matW}{W} 
\newcommand{\matS}{S} 
\newcommand{\matM}{M} 
 
\newcommand{\Xvecbar}{\overline{{\bf x}}}
\newcommand{\vecmu}{\mbox{\boldmath $\mu$}}
\newcommand{\matB}{B} 
\newcommand{\tr}{\mbox{\it tr}} 
\newcommand{\amu}{\alpha_{\mu}} 
\newcommand{\aw}{\alpha_W} 
\newcommand{\awY}{\alpha_{W,\bY}}

\title{Likelihoods and Parameter Priors for Bayesian Networks}

\comment{
\author{
David Heckerman and Dan Geiger\thanks{\ Author's primary affiliation:
Computer Science Department, Technion, Haifa 32000, Israel.}\\
\\
Microsoft Research, Bldg 9S/1\\
Redmond 98052-6399, WA\\
heckerma@microsoft.com, dang@cs.technion.ac.il}

\date{}
}

\author{
David Heckerman\\
heckerma@hotmail.com\\
\\
Dan Geiger\\ 
geiger02@gmail.com}

\date{November 1995, Revised June 2021}

\begin{document}

\maketitle

\begin{abstract}

\noindent We develop simple methods for constructing likelihoods
and parameter priors for learning about the parameters and structure
of a Bayesian network.  In particular, we introduce several
assumptions that permit the construction of likelihoods and parameter
priors for a large number of Bayesian-network structures from a small
set of assessments.  The most notable assumption is that of likelihood
equivalence, which says that data can not help to discriminate network
structures that encode the same assertions of conditional
independence.  We describe the constructions that follow from these
assumptions, and also present a method for directly computing the
marginal likelihood of a random sample with no missing observations.
Also, we show how these assumptions lead to a general framework for
characterizing parameter priors of multivariate distributions.

\bigskip


\bigskip

\noindent
Corrections to the original text in \textcolor{red}{red} are taken
from J. Kuipers, G. Moffa, and D. Heckerman, Addendum on the scoring
of Gaussian directed acyclic graphical models. {\em Annals of Statistics} 
42, 1689-1691, Aug 2014. 
Other updates to the original are in \textcolor{blue}{blue}.

\end{abstract}

\section{Introduction} \label{sec:intro}

A Bayesian network is a graphical representation of a probabilistic
model that most people find easy to construct and interpret (Howard,
1981; Pearl, 1988).\nocite{Howard81,Pearl88} Over the last decade, the
Bayesian network has become a popular representation for encoding
uncertain expert knowledge in expert systems (Heckerman et al.,
1995a).\nocite{HMW95cacm} More recently, researchers have developed
methods for learning Bayesian networks from a combination of expert
knowledge and data.  In this paper, we describe methods for assigning
likelihoods and parameter priors for this learning problem.

Let $\U=\{\X_1,\ldots,\X_n\}$ be a set of random variables.  We use
$\x_i$ to denote a value of $\X_i$ and $\by=(\x_i)_{\X_i \in \bY}$ to
denote a value of $\bY \subseteq \U$.  A Bayesian network for $\U$ is
a graphical factorization of the joint probability distribution of
$\U$.  The representation consists of two components: a structure and
a set of local probability distributions.  The structure $\Bs$ for
$\U$ is a directed acyclic graph that represents a set of
conditional-independence assertions about the variables $\U$.  Namely,
let $(\X_1,\ldots,\X_n)$ be an ordering of $\X$.  For every variable
$\X_i$ in $\U$, there is a corresponding node $\X_i$ in $\Bs$, and a
set $\PaSi \subset \{\X_1,\ldots,\X_{i-1}\}$ corresponding to the
parents of $\X_i$ in $\Bs$.\footnote{We use the same symbol for a
variable and its corresponding node.}  The structure $\Bs$ for $\U$
represents the assertions that, for $i=2,\ldots,n$, $\X_i$ and
$\{\X_1,\ldots,\X_{i-1}\} \setminus \PaSi$ are independent given
$\PaSi$.  That is,
\begin{equation} \label{eq:bn-def}
p(\C) = \prod_{i=1}^n p(\x_i|\paSi)
\end{equation}
The local distributions associated with the Bayesian network are
precisely those in \Eq{eq:bn-def}.

In this discussion, we assume that the local distributions depend on a
finite set of parameters $\TBs \in \CTBs$.  Thus, we rewrite
\Eq{eq:bn-def} as follows:
\begin{equation} \label{eq:bn-def-p-noh}
p(\C|\TBs) = \prod_{i=1}^n p(\x_i|\paSi,\Thi)
\end{equation}
where $\TBs = (\Th_1,\ldots,\Th_n)$.  We assume that $\CTBs$ is
absolutely continuous.

Let $\hBs$ denote the assertion or hypothesis that the joint
distribution of $\U$ can be factored according to the structure
$\Bs$.  That is, define $\hBs$ to be true if there exists $\TBs \in
\CTBs$, where $\CTBs$ is absolutely continuous, such that
\Eq{eq:bn-def-p-noh} holds.  It will be useful to include this
hypothesis explicitly in the factorization of the joint distribution.
In particular, we write
\begin{equation} \label{eq:bn-def-p}
p(\C|\TBs,\hBs) = \prod_{i=1}^n p(\x_i|\pai,\Thi,\hBs)
\end{equation}
This notation often makes it unnecessary to use the superscript $\Bs$
in the term $\paSi$, and we use the simpler expression where
possible.

Let us consider the situation where both the parameters $\TBs$ and the
structure hypothesis $\hBs$ are uncertain.  Given data
$\Db=\{\C_1,\ldots,\C_m\}$, a random sample from $p(\U|\TBs,\hBs)$
where $\TBs$ and $\hBs$ are the true parameters and structure
hypothesis, respectively, we can compute the posterior probability of
an arbitrary structure hypotheses $\hBs$ using
\begin{equation} \label{eq:post-bs}
p(\hBs|\Db) = c \ p(\hBs) \ p(\Db|\hBs) 
  = c \ p(\hBs) \int p(\Db|\TBs,\hBs) \ p (\TBs|\hBs) \ d \TBs
\end{equation}
where $c$ is a normalization constant.  We can then select a model
(i.e., structure) that has a high posterior probability or average
several good models for prediction.  Methods for searching through the
space of Bayesian-network structures are discussed by Cooper and
Herskovits (1992), Aliferis and Cooper (1994), and Heckerman et al.
(1995b).\nocite{CH92,AC94uai,HGC95ml}

A difficulty with this approach arises when many network structures
are possible.  In this case, we need to assign likelihoods, structure
priors, and parameter priors to a large number of (if not all
possible) structures to enable a search among these models.  Buntine
(1991) and Heckerman et al. (1995b) discuss methods for determining
structure priors from a small number of direct assessments.  In this
paper, we develop practical methods for assigning likelihoods and
parameter priors to a large number of structures.  In particular, we
describe a set of assumptions under which likelihoods and parameter
priors can be determined by a relatively small number of direct
assessments.  We show how likelihoods and priors are constructed and
how marginal likelihoods $p(\Db|\hBs)$ are computed from these
assessments.  Some of our assumptions are abstracted from those made
previously by researchers who examined cases where the local
likelihoods are unrestricted discrete distributions (Cooper and
Herskovits, 1992; Spiegelhalter et al., 1993; Heckerman et al.,
1995b)\nocite{CH92,SDLC93,HGC95ml} and linear-regression models (Geiger
and Heckerman, 1994; Heckerman and Geiger,
1995)\nocite{GH94uai,HG95uai}.  The most notable assumptions are
global parameter independence, which says that the parameter variables
$\CTh_1,\ldots,\CTh_n$ are mutually independent, and likelihood
equivalence, which (roughly speaking) says that data can not help to
discriminate structures that encode the same assertions of conditional
independence.

An important outgrowth of our work is a framework for characterizing
prior distributions for the parameters of multivariate distributions.
In particular, for a given family of local likelihoods
$p(\x_i|\pai,\Thi,\hBs)$, the application of parameter independence
and likelihood equivalence yields a functional equation, the solution
to which delimits all allowed prior distributions under these
assumptions.  For the likelihoods that we have studied, these
solutions correspond to well-known distribution families.  Namely,
when likelihoods are unrestricted discrete distributions, the solution
to the functional equation is the Dirichlet distribution.  When $\X$
contains two variables and likelihoods are linear-regression models,
the only solution to the functional equation is the bivariate
normal-Wishart distribution.

\section{Examples} \label{sec:eg}

We illustrate the ideas in this paper using two standard
probability distributions, which we review in this section.
In the first case, each variable $\X_i \in \U$ is discrete, having
$r_i$ possible values $\x^1_i,\ldots,\x^{r_i}_i$.\footnote{When we
refer to an arbitrary value of $\X_i$, we drop the superscript.}
Each local likelihood is an unrestricted discrete distribution
\begin{equation} \label{eq:disc-i}
p(\x^k_i|\pai^j,\Thi,\hBs) = \theta_{x^k_i|\pai^j} \equiv \tijk
\end{equation}
where $\pai^1,\ldots,\pai^{q_i}$ ($q_i = \prod_{\X_i
\in \Pai} r_i$) denote the values of $\Pai$.
The local parameters are given by $\Thi=
((\tijk)_{k=1}^{r_i})_{j=1}^{q_i}$.  We assume that each parameter
$\tijk$ is greater than zero.

In the second case, each variable $\X_i \in \U$ is continuous, and
each local likelihood is the linear-regression model
\begin{equation} \label{eq:norm-i}
p(\x_i|\pai,\Thi,\hBs)=
  N(\x_i|m_i + \sum_{\x_j \in \pai} b_{ji} \x_j, 1/v_i)
\end{equation}
where $N(\x_i|\mu,\tau)$ is a normal distribution with mean $\mu$ and
precision $\tau>0$.  Given this form, a missing arc from $\X_j$ to
$\X_i$ implies that $b_{ji}=0$ in the full regression model.  The
local parameters are given by $\Thi=(m_i,\vecb_i,v_i)$, where
$\vecb_i$ is the column vector $(b_{1i}, \ldots, b_{i-1,i})$.  We call
a Bayesian network constructed with these likelihoods a {\em Gaussian
network} after Shachter and Kenley (1989).

\section{Simplifying Assumptions} \label{sec:ass}

In this section, we present assumptions that simplify the assessment
of likelihoods and parameter priors.  In this explication, we consider
situations where all structure hypotheses for $\X$ are possible---that
is, $p(\hBs)>0$ for all $\Bs$ for $\X$.

The first assumption, already mentioned, is that
$\CTh_1,\ldots,\CTh_n$ are mutually independent.

\begin{assumption}[Global Parameter Independence] \label{ass:pi}
Given any structure $\Bs$ for $\U$, 
\[
p(\TBs|\hBs) = \prod_{i=1}^n p(\Thi|\hBs)
\]
\end{assumption}

\noindent
Spiegelhalter and Lauritzen (1990)\nocite{SL90} introduced this
assumption in the context Bayesian networks under the name global
independence.

Roughly speaking, the next two assumptions capture the notion that the
likelihoods and priors are modular in the sense that these quantities
for variable $\X_i \in \U$ depend only the structure that is local to
$\X_i$---namely, the parents of $\X_i$---and not on the entire
structure.

\begin{assumption}[Likelihood Modularity] \label{ass:lm}
Given any structure $\Bs$ for $\U$, 
\begin{equation} \label{eq:lm}
p(\x_i|\paSi,\Thi,\hBs) = p(\x_i|\paSi,\Thi)
\end{equation}
for all $\X_i \in \U$.
\end{assumption}

\begin{assumption}[Prior Modularity] \label{ass:pm}
Given any two structures $\Bsone$ and $\Bstwo$ for
$\U$ such that $\X_i$ has the same parents in $\Bsone$ and $\Bstwo$,
\[
p(\Thi|\hBsone) = p(\Thi|\hBstwo)
\]
\end{assumption}

\noindent
Both assumptions have been used implicitly in the work of (e.g.)
Cooper and Herskovits (1992), Spiegelhalter et al. (1993), and Buntine
(1994).\nocite{CH92,SDLC93,Buntine94} Heckerman et
al. (1995b)\nocite{HGC95ml} made \Ass{ass:pm} explicit under the name
parameter modularity.

The assumption of likelihood modularity holds in our examples (see
Equations~\ref{eq:disc-i} and \ref{eq:norm-i}).  To illustrate the
assumption of prior modularity, consider the set of binary
variables $\U=\{X,Y\}$.  In both the structures $\Bxtoy$
($\xtoy$) and $\Bxy$ (no arc between $X$ and $Y$), the node $X$ has
the same parents (none).  Consequently, by prior modularity, we
have that $p(\Th_{x}|\hBxtoy) = p(\Th_{x}|\hBxy)$.

The next two assumptions relate to the notion of structure
equivalence.  Consider the two structures
\mbox{$\X_1 \rightarrow \X_2 \rightarrow \X_3$}
and \mbox{$\X_1 \leftarrow \X_2 \leftarrow \X_3$}.  Both structures
represent the assertion that $\X_1$ and $\X_3$ are conditionally
independent given $\X_2$ and no other assertions of independence.  In
general, we say that two structures for $\U$ are {\em independence
equivalent} if they represent the same assertions of conditional
independence.  Independence equivalence is an equivalence relation,
and induces a set of equivalence classes over the possible structures
for $\X$.  Verma and Pearl (1990)\nocite{Verma90} provides a simple
characterization of independence-equivalent structures.  Given a
structure $\Bs$, a {\em v-structure} in $\Bs$ is an ordered node
triple $(\X_i,\X_j,\X_k)$ where $\Bs$ contains the arcs $\X_i
\rightarrow \X_j$ and $\X_j \leftarrow \X_k$, and there is no arc
between $\X_i$ and $\X_k$ in either direction.

\begin{theorem}[Verma and Pearl, 1990]\nocite{Verma90} \label{thm:VP90} 
Two structures for $\U$ are independence equivalent if and
only if they have identical edges and identical v-structures.
\end{theorem}

\noindent
This characterization makes it easy to identify independence
equivalent structures.  The following characterization by Chickering
(1995) is useful for proving technical claims about independence
equivalence.  An {\em arc reversal} is a transformation from one
structure to another, in which a single arc between two nodes is
reversed.  An arc between two nodes is said to be {\em covered} if
those two nodes would have the same parents if the arc were
removed.

\begin{theorem}[Chickering, 1995] \label{thm:C95}
Two structures for $\U$ are independence equivalent if and only if
there exists a set of covered arc reversals that transform one
structure into the other.
\end{theorem}

A concept related to that of independence equivalence is that of
distribution equivalence.  As is typically done is practice, we assume
that the local likelihoods $p(\x_i|\pai,\Thi,\hBs)$ are restricted to
some family of probability distributions ${\cal F}$.  Then, $\Bsone$
and $\Bstwo$ are {\em distribution equivalent with respect to (wrt)
${\cal F}$} if the two structures represent the same set of
distributions---that is, for every $\TBsone$, there exists a $\TBstwo$
such that $p(\C|\TBsone,\hBsone)=p(\C|\TBstwo,\hBstwo)$, and vice
versa.

Distribution equivalence wrt some ${\cal F}$ implies independence
equivalence, but the converse does not hold.  Nonetheless, if all
structures that differ by a single arc reversal are distribution
equivalent wrt ${\cal F}$, then by
\Thm{thm:C95}, independence equivalence implies distribution
equivalence.  We adopt this assumption formally as follows.

\begin{assumption}[Covered-Arc-Reversal Equivalence] \label{ass:car-eq}
Given local likelihoods restricted to ${\cal F}$, any two
structures for $\U$ that differ by a single covered arc
reversal are distribution equivalent wrt ${\cal F}$.
\end{assumption}

\Ass{ass:car-eq} holds trivially in the discrete case.
Shachter and Kenley (1989) show that \Ass{ass:car-eq} holds in the
linear-regression case.  A case where \Ass{ass:car-eq} does not hold
is one where $\U$ consists of three or more binary variables and the
local likelihoods are restricted to the sigmoid function
\begin{displaymath}
p(\x_i|\pai,\Thi,\hBs)= \frac{1}{
  1 + {\rm exp}\left\{ a_i + \sum_{\x_j \in \pai} b_{ji} \x_j \right\}}
\end{displaymath}
where $\Thi=(a_i,\vecb_i)$.  For example, with
$\U=\{\X_1,\X_2,\X_3\}$, suppose $\Bsone$ is the structure with arcs
$\X_1 \rightarrow \X_2$, $\X_1 \rightarrow \X_3$, and $\X_2
\rightarrow \X_3$, and $\Bstwo$ is the structure with arcs $\X_1
\rightarrow \X_2$, $\X_1 \rightarrow \X_3$, and $\X_3
\rightarrow \X_2$.  Then, $\Bsone$ and $\Bstwo$ differ by the
reversal of a covered arc between $\X_2$ and $\X_3$, but, given the
sigmoid restriction, there are certain joint likelihoods that can
be represented by one structure, but not the other.

\comment{
An example where
\Ass{ass:car-eq} does not hold is the case where $\U$ contains
discrete variables, each variable has three or more values,
and the likelihood has the log-linear form
\[
\log p(\x_i|\pai,\Thi,\hBs)= \alpha_{\pai} + \beta_{\pai} \cdot \x_i
\]
where can take on different values for every value of $\Pai$.

Yet another example is the exponential likelihood.
}

We assume that the parameters $\Thi$ are uniquely determined given the
local likelihood $p(\x_i|\paSi,\Thi)$ (i.e., all parameters are
identified).  Consequently, given two structures $\Bsone$ and $\Bstwo$
that are distribution-equivalent wrt ${\cal F}$, for every $\TBsone$,
there exists a unique $\TBstwo$ such that
$p(\C|\TBsone,\hBsone)=p(\C|\TBstwo,\hBstwo)$.  That is, there is a
one-to-one mapping from $\TBsone$ to $\TBstwo$, which we write
$\TBstwo = \CTBstwo(\TBsone)$.  Also, we assume that, the Jacobian
$|\dXdY{\TBsone}{\TBstwo}|$ exists and is non-zero for all values of
$\CTBsone$.  These technical assumptions hold for our examples.

Given \Ass{ass:car-eq} and these technical assumptions, we can make
the following assumption.

\begin{assumption}[Marginal Likelihood Equivalence] \label{ass:le}
Given any two independence-equivalent structures $\Bsone$ and
$\Bstwo$ for $\U$, 
\[
p(\C|\CTBstwo(\TBsone),\hBstwo) = p(\C|\TBsone,\hBsone)
\]
and
\[ 
p(\TBstwo|\hBstwo) =  \left| \frac{\partial \TBsone}{\partial \TBstwo}
  \right| \ p(\TBsone|\hBsone)
\]
for all values of $\TBsone$.
\end{assumption}

\noindent
An immediate consequence of this assumption is that, given any two
independence-equivalent structure $\Bsone$ and $\Bstwo$,
$p(\Db|\hBsone)=p(\Db|\hBstwo)$---hence, the name ``marginal
likelihood equivalence.''  For the sake of brevity, we often refer to
this assumption as likelihood equivalence.

Given our definition of $\hBs$ in the introduction, likelihood
equivalence follows from covered-arc-reversal equivalence.  In
particular, by this definition, $\hBs$ is nothing more than a
constraint on the possible joint likelihoods.  Furthermore, given
\Ass{ass:car-eq}, whenever $\Bsone$ and $\Bstwo$ are independence
equivalent, $\hBsone$ and $\hBstwo$ correspond to the same constraint
on the possible joint likelihoods.  Thus, if $\hBsone$ and $\hBstwo$
are independence equivalent, then $\hBsone=\hBstwo$.  This property,
which we call {\em hypothesis equivalence,} implies likelihood
equivalence.

Nonetheless, some researchers give Bayesian-network structure a causal
interpretation (e.g., Spirtes et al., 1993; Pearl,
1995)\nocite{Spirtes93,Pearl95bm}.  In this case, we can modify the
definition of $\hBs$ to include the assertion that if $\X_i
\rightarrow \X_j$ in $\Bs$, then $\X_i$ is a direct cause of $\X_j$.
Consequently, hypothesis equivalence does not hold.  Nonetheless, the
weaker assumption of likelihood equivalence is sometimes reasonable.
For a detailed discussion of this point, see Heckerman
(1995)\nocite{H95uai}.  To allow for the causal interpretation of
structure, we take likelihood equivalence to be an assumption.

We close this section with a few observations about the acausal
interpretation of structure where hypothesis equivalence holds.  Given
this property, we can think of a model as an equivalence class of
structures rather than an individual structure.  Thus, for example, we
can search for good models by searching through the space of equivalence
classes.  Spirtes and Meek (1995)\nocite{SM95kdd} and Chickering
(1995)\nocite{C95tr} describe such search methods.  Also,
when we make a prediction by averaging models, we can average
over equivalence classes.

Model averaging raises another important point.  To average model
predictions, the structure hypotheses should be mutually exclusive.
Without the assumption that parameters are absolutely continuous,
however, hypotheses are not mutually exclusive.  For example, given
$\U=\{X,Y\}$, both structure hypotheses $\hBxtoy$ and $\hBxy$ include
the case where $X$ and $Y$ are independent.  In fact, given $\hBxtoy$,
if we assign non-zero priors only to those parameter values that
encode independence between $X$ and $Y$, then $\hBxtoy$ implies
$\hBxy$.  Fortunately, our technical assumption implies that, if
$\Bsone$ and $\Bstwo$ are not independence equivalent, then $\hBsone$
and $\hBstwo$ are mutually exclusive, that is $p(\hBstwo,\hBsone)=0$.

In particular, assume that $\Bsone$ and $\Bstwo$ are not independence
equivalent.  Given the definition of structure hypothesis, we have that
\begin{displaymath}
p(\hBstwo|\hBsone,\TBsone) = \left\{
  \begin{array}{ll}
    1 & \exists \TBstwo {\rm \ s.t. \ } 
        p(\C|\TBstwo,\hBstwo)=p(\C|\TBsone,\hBsone) \\
    0 & {\rm otherwise}
  \end{array} \right.
\end{displaymath}
If $p(\hBstwo|\hBsone,\TBsone)=0$ almost everywhere in
$\CTBsone$, then, because $\CTBsone$ is absolutely continuous,
\begin{displaymath}
p(\hBstwo|\hBsone) = 
  \int p(\hBstwo|\TBsone,\hBsone) \ p(\TBsone|\hBsone) \ d \TBsone
  = 0
\end{displaymath}
Thus, mutually exclusivity holds because, by assumption, $p(\hBsone)>0$.
If $p(\hBstwo|\hBsone,\TBsone) \neq 0$ almost everywhere in
$\CTBsone$, then the independencies encoded by $\Bstwo$ must be a
proper subset of those encoded by $\Bsone$.  (E.g., $\Bsone=\Bxy$ and
$\Bstwo=\Bxtoy$.)  In this case, given $\hBsone$, the parameters
$\CTBstwo$ can not be absolutely continuous, and
$p(\hBstwo|\hBsone)=0$.

\section{The Construction of Likelihoods Parameter Priors} 
\label{sec:like-pri}

Given Assumptions~\ref{ass:pi} through \ref{ass:le} and the technical
assumptions,\footnote{To simplify the presentation, we omit explicit
reference to the technical assumptions in the remainder of the paper.}
we can construct the likelihoods and parameter priors for all
structures for $\U$ from a small number of assessments.  In this
section, we describe these constructions.  An important concept in
this approach is that of a complete structure: one that has no missing
arcs.  All complete structures for a given $\U$ are independence
equivalent.

First, let us consider likelihoods.

\begin{theorem} \label{thm:like}
Given Assumptions~\ref{ass:lm}, \ref{ass:car-eq}, and \ref{ass:le}, we
can construct the likelihood $p(\C|\TBs,\hBs)$ for any 
structure $\Bs$ given $p(\C|\TBsc,\hBsc)$ for any complete 
structure $\Bsc$.
\end{theorem}

\noindent {\bf Proof:} Given any structure $\Bs$ for $\U$,
we construct the likelihoods $p(\x_i|\pai,\Thi,\hBs)$ as follows.  For
each $i=1,\ldots,n$, we find a complete structure $\Bsci$ such
that $\PaSi=\PaScii$.  Then, we use likelihood equivalence to compute
$p(\C|\TBsci,\hBsci)$ and hence $p(\x_i|\pai,\Thi,\hBsci)$ from
$p(\C|\TBsc,\hBsc)$.  Using likelihood modularity we obtain
$p(\x_i|\pai,\Thi,\hBs)=p(\x_i|\pai,\Thi,\hBsci)$.
\qed

The construction of parameter priors is similar.

\begin{theorem} \label{thm:prior}
Given Assumptions~\ref{ass:pi}, \ref{ass:pm},
\ref{ass:car-eq}, and \ref{ass:le}, we can construct the prior 
$p(\TBs|\hBs)$ for any structure $\Bs$ given $p(\TBsc|\hBsc)$
for any complete structure $\Bsc$.
\end{theorem}

\noindent {\bf Proof:} Given any structure $\Bs$ for $\U$,
we determine $p(\TBs|\hBs)$ by computing $p(\Thi|\hBs)$, $i=1,\ldots,n$,
and then applying global parameter independence.  To determine
$p(\Thi|\hBs)$, we identify a complete structure such that
$\PaSi=\PaScii$.  Then, we use likelihood equivalence to
compute $p(\TBsci|\hBsci)$ from $p(\TBsc|\hBsc)$.  Next, we apply
global parameter independence to obtain $p(\Thi|\hBsci)$.  Finally, we
use prior modularity, which gives $p(\Thi|\hBs)=p(\Thi|\hBsci)$.
\qed

Given likelihood equivalence, we can compute $p(\C|\TBsc,\hBsc)$ and
$p(\TBsc|\hBsc)$ for one complete structure from the
likelihood and prior for another complete structure.  In so
doing, we are simply performing coordinate transformations between
parameters for different variable orderings in the factorization of
the joint likelihood.  Thus, likelihood equivalence raises the
possibility of defining a unique joint likelihood $p(\C|\ThU,\hBsc)$
whose parameters $\ThU$ are variable-order independent.  Given this
likelihood and the corresponding prior $p(\ThU|\hBsc)$, we can compute
$p(\C|\TBsc,\hBsc)$ and $p(\TBsc|\hBsc)$ for any complete 
structure and, using the techniques described previously in this
section, the likelihoods and parameters priors for any 
structure.  Note that the choice of $p(\C|\TBsc,\hBsc)$ must be
consistent with global parameter independence and likelihood
equivalence.  We address this issue in \Sec{sec:consis}.

Let us consider this approach for our discrete and linear-regression
examples.  In the discrete case, the joint likelihood is the
$n$-dimensional multivariate-discrete distribution:
\[
p(\x_1,\ldots,\x_n|\ThU,\hBsc) = \theta_{\x_1,\ldots,\x_n}
\]
where $\ThU = (\theta_{\x_1,\ldots,\x_n})_{(\x_1,\ldots,\x_n) \in \U}$.  The
one-to-one mapping from $\TBsc$ for the complete structure with
ordering $(\X_1,\ldots,\X_n)$ to $\ThU$ is given by
\begin{eqnarray} \label{eq:disc-thu}
\theta_{\x_1,\ldots,\x_n} = \prod_{i=1}^n \theta_{\x_i|\x_1,\ldots,\x_{i-1}}
\end{eqnarray}
The Jacobian for this mapping (and its inverse) exist and are non-zero
for all allowed values of $\TBsc$.  The Jacobian for the inverse
mapping, derived by Heckerman et al. (1995b)\nocite{HGC95ml} is given
by
\begin{equation} \label{eq:jac-d}
\left| \frac{\partial \ThU}{\partial \TBsc} \right| =
  \prod_{i=1}^{n-1} \prod_{x_1,\ldots,x_{i}} [
  \ta{x_i|x_1,\ldots,x_{i-1}} ]^{[\prod_{j=i+1}^n r_j] - 1}
\end{equation}

In the linear-regression example, the joint likelihood is 
the $n$-dimensional multivariate-normal distribution
with mean $\vecmu$ and symmetric positive definite 
precision matrix $\matW$
\[
p(\C|\ThU,\hBsc) = {\bf N}_n(\C|\vecmu, \matW)
\]
The one-to-one mapping from $\TBsc$ for the complete structure
with ordering $(\X_1,\ldots,\X_n)$ to $\ThU=\{\vecmu,\matW\}$ is given
by
\begin{equation} \label{eq:m-mu}
\mu_i = m_i + \sum_{j=1}^{i-1} b_{ji} \mu_j
\end{equation}
and the recursive formula
\begin{eqnarray} \label{eq:shachter} 
\matW(1) & = & \frac{1}{v_1} \\
\matW(i+1) & = & \left( \begin{array}{cc} 
\matW(i) + \frac{\vecb_{i+1} \vecb'_{i+1}}{v_{i+1}} & 
-\frac{\vecb_{i+1}}{v_{i+1}}   \\ 
-\frac{\vecb'_{i+1}}{v_{i+1}} &  \frac{1}{v_{i+1}} 
\end{array} \right), \ \ \ i>1 \nonumber
\end{eqnarray}
where $W(i)$ is the $i \times i$ upper left submatrix of $\matW$
(e.g., Shachter and Kenley, 1989\nocite{Shachter89b}), 
\textcolor{blue}{and the prime symbol denotes transpose.}
The Jacobian
for this mapping (and its inverse) exist and are non-zero for all
allowed values of $\TBsc$.  Let $\vecm=(m_1,\ldots,m_n)$,
$\vecv=(v_1,\ldots,v_n)$ and $\matB=(\vecb_1,\ldots,\vecb_n)$.  The
Jacobian for the mapping from $\vecmu$ to $\vecm$ for a given $\matB$
is
\begin{equation}  \label{eq:jac-m} 
\left| \frac{\partial \vecmu}{\partial \vecm} \right| = 1
\end{equation}
The Jacobian for the mapping from $\matW$ to $(\vecv,\matB)$
is easily obtained from \Eq{eq:shachter}:
\begin{equation}  \label{eq:jac-w} 
\left| \frac{\partial \matW}{\partial \vecv \matB} \right| =
  \prod_{i=1}^n v_i^{-(i+1)} 
\end{equation} 

To illustrate our techniques for constructing priors, consider again
the simple two-binary-variable case $\U=\{X,Y\}$.  Suppose
$p(\ThU|\hBxtoy)$ is the Dirichlet distribution
\[
p(\ThU|\hBxtoy) = c \ \prod_{xy} \ta{xy}^{\alpha \cdot p(xy|\hBxtoy)}
\]
where $\alpha$ is an 
\textcolor{blue}{effective} sample size and $p(xy|\hBxtoy) = \int
\theta_{xy} p(\ThU|\hBxtoy) d \ThU$.  That is, $p(xy|\hBxtoy)$ is
the marginal likelihood that $X=x$ and $Y=y$ in a one-sample random
sample.  Given this prior, we compute $p(\TBxy|\hBxy)$ for the
structure containing no arc between $X$ and $Y$ as follows.  First,
we use Equations~\ref{eq:disc-thu} and \ref{eq:jac-d} to change
variables to $\TBxtoy$ obtaining
\[
p(\TBxtoy|\hBxtoy) = c \   
  \prod_x \ta{x}^{\alpha \cdot p(x|\hBxtoy)} \ 
  \prod_{xy} \ta{y|x}^{\alpha \cdot p(y|x,\hBxtoy)}
\]
Note that this prior satisfies global parameter independence.
Integrating out all parameters except $\ta{x}$ and using prior
modularity, we get
\begin{equation} \label{eq:tax}
p(\ta{x}|\hBxy) = c \ \prod_x \ta{x}^{\alpha \cdot p(x|\hBxtoy)}
\end{equation}
Likewise, changing variables to $\TBytox$, integrating, and applying
prior modularity, we have
\begin{equation} \label{eq:tay}
p(\ta{y}|\hBxy) = c \ \prod_y \ta{y}^{\alpha \cdot p(y|\hBxtoy)}
\end{equation}
Equations~\ref{eq:tax} and \ref{eq:tay} together with global
parameter independence yield the desired prior.

\section{Computation of the Marginal Likelihood for Complete Data} 
\label{sec:be}

Given a structure $\Bs$, we can use the techniques developed in the
last section to construct the likelihood and parameter prior, and then
apply \Eq{eq:post-bs} to compute the marginal likelihood $p(D|\hBs)$.
In this section, we present a short cut to this approach for
situations where the random sample $\Db$ is complete (i.e., contains
no missing observations).  In particular, we derive a formula for the
marginal likelihood of a complete random sample that bypasses the
explicit construction of likelihoods and priors.

For a given $\U$, consider any structure $\Bs$ and complete random
sample $\Db$.  Assuming global parameter independence, the parameters
remain independent given complete data.  That is,
\begin{equation} \label{eq:post-pi} 
p(\TBs|\Db,\hBs) = 
  \prod_{i=1}^n p(\Thi|\Db,\hBs) 
\end{equation}
In addition, assuming global parameter independence, likelihood
modularity, and prior modularity, the parameters remain modular
given complete data.  In particular, if $\X_i$ has the same parents in
$\Bsone$ and $\Bstwo$, then 
\begin{equation} \label{eq:post-pm} 
p(\Thi|\Db,\hBsone) = p(\Thi|\Db,\hBstwo)
\end{equation}
Also, for any $\bY \subseteq \U$, define $\Db^{\bY}$ to be the random
sample $\Db$ restricted to observations of $\bY$.  For example, if
$\U=\{\X_1,\X_2,\X_3\}$, $\bY=\{\X_1,\X_2\}$, and
$\Db=\{\C_1=\{\x_{11},\x_{12},\x_{13}\},
\C_2=\{\x_{21},\x_{22},\x_{23}\}\}$, then we have $\Db^{\bY} = \{
\{\x_{11},\x_{12}\}, \{\x_{21},\x_{22}\}\}$.  
Let $\bY$ be a subset of $\U$, and $\Bsc$ be a complete structure for
any ordering where the variables in $\bY$ come first.  Then, assuming
global parameter independence and likelihood modularity, it is not
difficult to show that
\begin{equation} \label{eq:ignore}
p(\bY|\Db,\hBsc) = p(\bY|\Db^{\bY},\hBsc)
\end{equation}
Given these observations, we can compute the marginal likelihood as
follows.

\begin{theorem} \label{thm:be}
Given any complete structure $\Bsc$ for $\U$, any structure $\Bs$ for
$\U$, and any complete random sample $\Db$, Assumptions~\ref{ass:pi}
through \ref{ass:le} imply
\begin{equation} \label{eq:be}
p(\Db|\hBs) = \prod_{i=1}^n 
  \frac{p(\Db^{\Pai \cup \{\X_i\}}|\hBsc)}{
        p(\Db^{\Pai}|\hBsc)}
\end{equation}
\end{theorem}

\noindent {\bf Proof:}
From the rules of probability, we have

\begin{equation} \label{eq:be1a}
p(\Db|\hBs) = \prod_{l=1}^m \int p(\C_l|\TBs,\hBs) \ p(\TBs|\Db_l,\hBs) 
  \ d\TBs
\end{equation}

\comment{reverse terms
\begin{eqnarray*} \label{eq:be1a}
\lefteqn{ p(\Db|\hBs) = \prod_{l=1}^m \int p(\TBs|\Db_l,\hBs) } \\
&& \cdot p(\C_l|\TBs,\hBs) \ d\TBs
\end{eqnarray*}
}

\noindent
where $\Db_l = \{\C_1,\ldots,\C_{l-1}\}$.  Using
Equations~\ref{eq:bn-def-p} and \ref{eq:post-pi} to rewrite the first
and second terms in the integral, respectively, we obtain

\begin{displaymath} \label{eq:be1b}
p(\Db|\hBs) = \prod_{l=1}^m \int \prod_{i=1}^n p(\x_{il}|\pail,\Thi,\hBs) \ 
  p(\Thi|\Db_l,\hBs) \ d\TBs
\end{displaymath}

\noindent
Using likelihood modularity and \Eq{eq:post-pm}, we get

\begin{equation} \label{eq:be2}
p(\Db|\hBs) = \prod_{l=1}^m \int \prod_{i=1}^n p(\x_{il}|\pail,\Thi,\hBsci) \ 
  p(\Thi|\Db_l,\hBsci) \ d\TBs
\end{equation}

\comment{reverse terms
\begin{eqnarray*} \label{eq:be2}
\lefteqn{ p(\Db|\hBs) = \prod_{l=1}^m \int \prod_{i=1}^n 
  p(\Thi|\Db_l,\hBsci) } \\
&& \cdot p(\x_{il}|\pail,\Thi,\hBsci) \ d\TBs
\end{eqnarray*}
}

\noindent
where $\Bsci$ is a complete structure with variable ordering
$\Pai$, $\X_i$ followed by the remaining variables.  Decomposing
the integral over $\TBs$ into integrals over the individual parameter
sets $\Thi$, and performing the integrations, we have
\begin{displaymath} \label{eq:be3}
p(\Db|\hBs) = \prod_{l=1}^m \prod_{i=1}^n 
  p(\x_{il}|\pail,\Db_l,\hBsci)
\end{displaymath}
Using \Eq{eq:ignore}, we obtain
\begin{eqnarray} \label{eq:be4}
p(\Db|\hBs) & = & \prod_{l=1}^m \prod_{i=1}^n 
  \frac{ p(\x_{il},\pail|\Db_l,\hBsci) }{ p(\pail|\Db_l,\hBsci) }
     \nonumber \\*[9pt]
& = & \prod_{l=1}^m \prod_{i=1}^n 
  \frac{
     p(\x_{il},\pail|\Db^{\Pai \cup \{\X_i\}}_l,\hBsci) }{
     p(\pail|\Db^{\Pai}_l,\hBsci) }
     \nonumber \\*[9pt]
& = & \prod_{i=1}^n 
  \frac{
     p(\Db^{\Pai \cup \{\X_i\}}|\hBsci) }{
     p(\Db^{\Pai}|\hBsci) }
\end{eqnarray}
By likelihood modularity and likelihood equivalence, we have that
$p(\Db|\hBsci) = p(\Db|\hBsc), i=1,\ldots,n$.
Consequently, for any subset $\bY$ of $\U$, we obtain
$p(\Db^{\bY}|\hBsci) = p(\Db^{\bY}|\hBsc)$ by summing over
the variables in $D^{\U \setminus \bY}$.  Applying this result to
\Eq{eq:be4}, we get \Eq{eq:be}. \qed

To apply \Eq{eq:be}, we assume a prior for $p(\ThU|\hBsc)$ that is
consistent with global parameter independence and likelihood
equivalence, and use this prior to compute the individual terms in
\Eq{eq:be}.  In the remainder of this section, we illustrate this
approach for the discrete and linear-regression models.  We
demonstrate consistency in \Sec{sec:consis}.

For the discrete model, we assume that $p(\ThU|\hBsc)$ is the
Dirichlet distribution:
\begin{equation} \label{eq:joint-dir}
p(\ThU|\hBsc) =
  \prod_{\x_1,\ldots,\x_n}  
    \theta_{\x_1,\ldots,\x_n}^{
      \alpha \cdot p(\x_1,\ldots,\x_n|\hBsc) - 1}
\end{equation}
where $\alpha$ is an 
\textcolor{blue}{effective} sample size, and the probabilities are
defined as they were in our two-variable example.  It follows that,
for any $\bY \subseteq \U$, the parameter set $\Th_{\bY}$ also has a
Dirichlet distribution:
\begin{equation} \label{eq:joint-dirX}
p(\Th_{\bY}|\hBsc) =
  \prod_{\by}  
    \theta_{\by}^{
      \alpha \cdot p(\by|\hBsc)-1}
\end{equation}
(e.g., DeGroot, 1970, p.\ 50)\nocite{DeGroot70}.  Furthermore, the
marginal likelihood for $\Db^{\bY}$ is given by
\begin{equation} \label{eq:ml-dir}
p(\Db^{\bY}|\hBsc) = \frac{\Gamma(\alpha)}{\Gamma(\alpha + m)}
  \prod_{\by} \frac{\Gamma(\alpha \cdot p(\by|\hBsc) + N_{\by})}{
                    \Gamma(\alpha \cdot p(\by|\hBsc))}
\end{equation}
where $m$ is the number of samples in $\Db$, and $N_{\by}$ is the
number of samples in $\Db$ where $\bY=\by$.  Combining
Equations~\ref{eq:be} and \ref{eq:ml-dir}, we obtain

\begin{equation} \label{eq:bde}
p(\Db|\hBs) = \prod_{i=1}^n \prod_{j=1}^{q_i}
  \frac{\Gamma(\alpha_{ij})}{\Gamma(\alpha_{ij}+N_{ij})} 
  \prod_{k=1}^{r_i} \frac{\Gamma(\alpha_{ijk} +
  N_{ijk})}{\Gamma(\alpha_{ijk})}
\end{equation}

\comment{
\begin{eqnarray} \label{eq:bde}
p(\Db|\hBs) & = & \prod_{i=1}^n \prod_{j=1}^{q_i}
  \frac{\Gamma(\alpha_{ij})}{\Gamma(\alpha_{ij}+N_{ij})} \nonumber \\
  && \cdot \prod_{k=1}^{r_i} \frac{\Gamma(\alpha_{ijk} +
  N_{ijk})}{\Gamma(\alpha_{ijk})}
\end{eqnarray}
}

\noindent
where

\begin{equation} \label{eq:npijk}
\alpha_{ijk} = \alpha \cdot p(\x_i^k,\pai^j|\hBsc), \ \ \ \ \ \ 
\alpha_{ij} = \sum_{k=1}^{r_i} \alpha_{ijk} = \alpha \cdot p(\pai^j|\hBsc)
\end{equation}

\comment{
\[
\alpha_{ijk} = \alpha \cdot p(\x_i^k,\pai^j|\hBsc)
\]
\begin{equation} \label{eq:npijk}
\alpha_{ij} = \sum_{k=1}^{r_i} \alpha_{ijk} = \alpha \cdot p(\pai^j|\hBsc)
\end{equation}
}

\noindent
$N_{ijk}$ is the number of samples where $\X_i=\x_i^k$ and 
$\Pai=\pai^j$, and $N_{ij}=\sum_{k=1}^{r_i} N_{ijk}$.
\Eqs{eq:bde} and \ref{eq:npijk} were originally derived in
Heckerman et al. (1994).

For the linear-regression model, we assume that $p(\vecmu,\matW|\hBsc)$ is 
a normal-Wishart distribution.  In particular, we assume that 
$p(\vecmu|\matW,\hBsc)$ is a multivariate-normal distribution
with mean $\vecmu_0$ and precision matrix $\amu
\matW$ ($\amu > 0$); and that $p(\matW|\hBsc)$ is a Wishart
distribution with $\aw$ degrees of freedom $(\aw > n-1)$ and
precision matrix
\textcolor{blue}{$\matT_0$}.
Thus, the posterior $p(\vecmu,\matW|\Db,\hBsc)$ is also a normal-Wishart
distribution.  In particular, 
$p(\vecmu|\matW,\Db,\hBsc)$ is multivariate normal with mean vector
$\vecmu_m$ given by
\begin{equation} 
\label{eq:updatemeans} 
\vecmu_m = \frac{\amu \vecmu_0 + m \Xvecbar_m}{\amu+m}
\end{equation} 
and precision matrix $(\amu+m)\matW$, where $\Xvecbar_m$ is the sample
mean of $\Db$; and $p(\matW|\Db,\hBsc)$ is a Wishart distribution with
$\aw+m$ degrees of freedom and matrix \textcolor{blue}{$\matT_m$} given by
\begin{equation} \label{eq:updateTau} 
\textcolor{blue}{\matT_m} =  
\textcolor{blue}{\matT_0} + \matS_m + \frac{\amu m}{\amu+m}
  (\vecmu_0-\Xvecbar_m)(\vecmu_0-\Xvecbar_m)' 
\end{equation} 
where $\matS_m$ is the \textcolor{red}{scatter matrix,} 
\textcolor{blue}{given by $\sum_{l=1}^m (\C_l - \Xvecbar_m)(\C_l - \Xvecbar_m)'$}
(e.g.,
DeGroot, 1970, p.\ 178).\nocite{DeGroot70} 
From these equations, we see
that $\amu$ and $\aw$ can be thought of as 
\textcolor{blue}{effective sample sizes for
the normal and Wishart components of the prior, respectively.}

Given $\bY \subseteq \U$ ($|Y|=l$), and vector ${\bf
z}=(z_1,\ldots,z_n)$, let ${\bf z}^{\bY}$ denote the vector formed by
the components $z_i$ of ${\bf z}$ such that $\X_i \in \bY$.
\textcolor{blue}{Similarly, given matrix $\matM$, let $\matM^{\bY\bY}$ denote the
submatrix of $\matM$ containing elements $m_{ij}$ such that $\X_i,\X_j
\in \bY$.}
If 
$p(\vecmu,\matW|\hBsc)$ is a normal-Wishart distribution as we have
described, then $p(\vecmu^{\bY},((\matW^-1)^{\bY\bY})^-1|\hBsc)$
is also a normal--Wishart distribution with constants $\vecmu_0^{\bY}$,
$\amu$, \textcolor{red}{$\matT_0^{\bY\bY}$}, and $\awY=\aw-n+l$
(e.g., see Press, 1971, Theorems 5.1.3 and 5.1.4).  Thus, 
we obtain the terms in \Eq{eq:be}:

\begin{equation} \label{eq:bge}
p(\Db^{\bY}|\hBsc)  = 
  \pi^{-lm/2} \left(\frac{\amu}{\amu+m}\right)^{l/2}
  \frac{c(l,\awY+m)}{c(l,\awY)} 
  |\textcolor{red}{\matT_0^{\bY\bY}}|^{\frac{\awY}{2}} 
  |\textcolor{red}{\matT_m^{\bY\bY}}|^{-\frac{\awY+m}{2}} 
  \nonumber
\end{equation}

\comment{
\begin{eqnarray} \label{eq:bge}
p(\Db^{\bY}|\hBsc) & = &
  \pi^{-lm/2} \left(\frac{\amu}{\amu+m}\right)^{l/2} \\
&& \cdot
  \frac{c(l,\awY+m)}{c(l,\awY)} 
  |\matW_0^{\bY\bY}|^{\frac{\awY}{2}} 
  |\matW_m^{\bY\bY}|^{-\frac{\awY+m}{2}} 
  \nonumber
\end{eqnarray}
}

\noindent
where
\begin{equation} \label{eq:cln} 
c(l,\alpha) = \prod_{i=1}^l \Gamma\left(\frac{\alpha+1-i}{2}\right)
\end{equation} 

\noindent
\textcolor{blue}{(See Geiger and Heckerman, 1994, for a derivation
when $l=n$.)}

\section{Priors from a Prior Bayesian Network} \label{sec:prior}

Whether we construct priors explicitly or use the short cut described
in the previous section, we require $p(\ThU|\hBsc)$ to compute
marginal likelihoods.  In this section, we discuss the assessment of
this distribution for our example models. \textcolor{blue}{We describe
one of multiple alternatives that makes use of what we call a {\em prior
Bayesian network}.}

In the discrete case, we can assess $p(\ThU|\hBsc)$ by assessing (1)
$p(\C|\hBsc)$ and 
(2) the \textcolor{blue}{effective} sample size $\alpha$.  Methods
for assessing $\alpha$ are discussed in (e.g.)  Heckerman et
al. (1995b).\nocite{HGC95ml} To assess $p(\C|\hBsc)$, we construct a
Bayesian network for 
\textcolor{blue}{$\C$ given $\hBsc$}, a prior Bayesian network.
\textcolor{blue}{We can then
derive the $\alpha_{ijk}$ using Equation~{\ref{eq:npijk}}.}

\textcolor{blue}{
For the linear-regression case, we directly assess the
effective sample sizes $\amu$ and $\aw$, and indirectly
assess $\vecmu_0$ and $\matT_0$.
For the latter assessments, 
we start with the observation that when
$p(\vecmu,\matW|\hBsc)$ is normal--Wishart as we have described, then
then $p(\C|\hBsc)$ is a multivariate $t$ distribution with $\aw-n+1$
degrees of freedom, location vector $\vecmu_0$, and precision matrix
$\amu(\aw-n+1)/(\amu+1)T_0^{-1}$. 
This result can be derived by first integrating over $\vecmu$ using
Equation 6 on p.\ 178 of DeGroot with sample size equal to one, and then
integrating over $\matW$
following an approach similar to that on pp.\ 179--180. Next, when
$\aw > n+1$, it follows that
\begin{equation} \label{eq:t1}
{\rm E}(\C|\hBsc) = \vecmu_0 \ \ \ \ \ \ 
{\rm Cov}(\C|\hBsc) = \frac{\amu+1}{\amu} \ \frac{1}{\aw-n-1} \ T_0
\end{equation}
(e.g., DeGroot, 1970, pp.\ 61).  
Thus, a person can assess a prior linear-regression Bayesian network 
for E$(\C|\hBsc)$ and Cov$(\C|\hBsc)$, and
then compute $\vecmu_0$ and $T_0$ using Equations~\ref{eq:t1}.
}

In both cases, the unusual aspect of this assessment is the
conditioning hypothesis $\hBsc$ (see Heckerman et al. [1995b] for a
discussion).

\comment{

For the linear-regression case, we propose an approach that is less
formal than the discrete case.  Here, we need to assess the
vector $\vecmu_0$, the matrix $\matT_0$, and the quantities $\amu>0$,
$\aw>n-1$ is as follows.  From Equations~\ref{eq:updatemeans} and
\ref{eq:updateTau} and the surrounding text, we see that $\mu_0$ and
$\matT_0$ can be thought of as an initial mean and scatter matrix,
respectively, and that $\amu$ and $\aw$ can be thought of as
effective sample sizes for them. Consequently, we can assess the
effective sample sizes directly, and then assess a prior Bayesian
network for the linear-regression model, infer its mean and
covariance, $\vecmu'_0$ and ${\rm Cov}(\U)$, respectively, and set

\begin{equation} \label{eq:t2}
\vecmu_0 = \vecmu_0' \ \ \ \ \ \ \matT_0 = \aw {\rm Cov}(\U).
\end{equation} 
}

\section{Consistency of the Dirichlet and Normal-Wishart Assumptions} 
\label{sec:consis}

In this section, we show that the Dirichlet and normal-Wishart priors
$p(\ThU|\hBsc)$ are consistent with the assumptions of global
parameter independence and likelihood equivalence.  To see the
potential for inconsistency, consider again the construction of
parameter priors in our two-binary-variable example.  Using $x$ and
$\bar{x}$ as a shorthand for the values $\x^1$ and $\x^2$, and a
similar shorthand for the values of $Y$, suppose we choose the prior

\[
p(\ThU|\hBxtoy)=
  p(\ta{xy}, \ta{\bar{x}y}, \ta{x\bar{y}}|\hBxtoy) 
  = \frac{c}{(\ta{xy} + \ta{x\bar{y}})  (1- (\ta{xy} + \ta{x\bar{y}}))} 
  = \frac{c}{\ta{x} (1 - \ta{x})}
\]

\comment{
\[
p(\ThU|\hBxtoy)=
  \frac{c}{(\ta{xy} + \ta{x\bar{y}})  (1- (\ta{xy} + \ta{x\bar{y}}))} 
\]
\vspace{-\medskipamount}
\[
\ \ \  = \frac{c}{\ta{x} (1 - \ta{x})}
\]
}

\noindent
Using likelihood equivalence and Equations~\ref{eq:disc-thu} and
\ref{eq:jac-d}, we obtain

\[
p(\ta{y}, \ta{x|y}, \ta{x|\bar{y}}|\hBytox) = 
  \frac{c \cdot \ta{y}(1-\ta{y})}{\ta{x} (1 - \ta{x})} = 
  \frac{c \cdot \ta{y}(1-\ta{y})}{
    (\ta{y}\ta{x|y}+(1-\ta{y})\ta{x|\bar{y}})
    (1 - (\ta{y}\ta{x|y}+(1-\ta{y})\ta{x|\bar{y}}))}
\]

\comment{
\[
p(\ta{y}, \ta{x|y}, \ta{x|\bar{y}}|\hBytox) = 
  \frac{c \cdot \ta{y}(1-\ta{y})}{\ta{x} (1 - \ta{x})} = 
\]
\vspace{-\medskipamount}
\[
  \frac{c \cdot \ta{y}(1-\ta{y})}{
    (\ta{y}\ta{x|y}+(1-\ta{y})\ta{x|\bar{y}})
    (1 - (\ta{y}\ta{x|y}+(1-\ta{y})\ta{x|\bar{y}}))}
\]
}

\noindent
which does not satisfy global parameter independence.

When $p(\ThU|\hBsc)$ is Dirichlet for some complete structure $\Bsc$,
however, likelihood equivalence implies that global parameter
independence holds for all complete structures.  We demonstrated this
fact for our two-variable example in \Sec{sec:like-pri}.  Heckerman et
al. (1995b, Theorem 3) prove the general case, which we summarize
here.\footnote{Note that the theorem is stated in a way that
presupposes likelihood equivalence.}

\begin{theorem} \label{thm:consis}
If the parameters $\ThU$ have the Dirichlet distribution
\begin{equation} \label{eq:given}
p(\ThU) = 
c \cdot \prod_{x_1,\ldots,x_n} 
  [\theta_{x_1,\ldots, x_n}]^{\alpha_{x_1,\ldots,x_n}-1}
\end{equation} 
then, for any complete structure $\Bsc$ for $\U$, the
distribution $p(\TBsc)$ satisfies global and local parameter
independence.  In particular,
\begin{equation} \label{eq:phidir3a}
p(\TBsc)= 
c \cdot \prod_{i=1}^n \prod_{x_1,\ldots, x_i}
[\theta_{x_i | x_1,\ldots, x_{i-1}}]^ {\alpha_{x_i|x_1,\ldots,x_{i-1}}-1}
\end{equation} 
where $\alpha_{x_i|x_1,\ldots,x_{i-1}} = \sum_{x_{i+1},\ldots,x_n}
\alpha_{x_1,\ldots,x_n}$.
\end{theorem}

\noindent {\bf Proof:}  The result follows by
multiplying the right-hand-side of \Eq{eq:given} by the
Jacobian \Eq{eq:jac-d}, using the relation
$\theta_{x_1,\ldots,x_n}= \prod_{i=1}^n
\theta_{x_i| x_1,\ldots,x_{i-1}}$, and collecting
powers of $\ta{x_i|x_1,\ldots,x_{i-1}}$. \qed

Consistency for the linear-regression case is shown in the
next theorem.

\begin{theorem} \label{thm:pinw} 
If $(\vecmu,\matW)$ has a normal--Wishart distribution, then
\[ 
p(\vecm, \vecv, \matB) = \prod_{i=1}^n p(m_i, v_i, \vecb_i) 
\] 
\end{theorem} 
 
\noindent {\bf Proof:}  To prove the theorem, we factor
$p(\vecm|\vecv,\matB)$ and $p(\vecv,\matB)$ separately.  By
assumption, we know that $p(\vecmu|\matW)$ is a multivariate-normal
distribution with mean $\vecmu_0$ and precision matrix $\amu W$.
Transforming this result to local distributions for $\mu_i$, we
obtain

\begin{equation}
 p(\mu_i|\mu_1,\ldots,\mu_{i-1},\vecv,\matB) = 
  \left(\frac{\amu}{2\pi v_i}\right)^{1/2} 
    \exp \left\{ 
    \frac{\left(\mu_i - \mu_{0i} - \sum_{j=1}^{i-1} b_{ji} 
      (\mu_j - \mu_{0j}) \right)^2}{2v_i/\amu}
  \right\}
\end{equation}

\comment{
\begin{eqnarray*}
\lefteqn{ p(\mu_i|\mu_1,\ldots,\mu_{i-1},\vecv,\matB) = 
  \left(\frac{\amu}{2\pi v_i}\right)^{1/2} } \\
&& \cdot \exp \left\{ 
    \frac{\left(\mu_i - \mu_{0i} - \sum_{j=1}^{i-1} b_{ji} 
      (\mu_j - \mu_{0j}) \right)^2}{2v_i/\amu}
  \right\}
\end{eqnarray*}
}

\noindent
for $i=1,\ldots,n$.
Using $m_{0i} = \mu_{0i} - \sum_{j=1}^{i-1} b_{ji} \mu_{0j}$,
collecting terms for each $i$, and using \Eq{eq:jac-m}, we have
\begin{equation} \label{eq:pinw1}
p(\vecm|\vecv,\matB) = \prod_{i=1}^n N(m_i|m_{0i},\amu/v_i)
\end{equation}
In addition, by assumption, $\matW$ has the Wishart distribution
\begin{equation} \label{eq:p-ind-w1}
p(\matW) =  
  c |\matW|^{(\aw-n-1)/2} e^{-1/2\tr\{\textcolor{blue}{\matT} \matW\}}  
\end{equation}
From \Eq{eq:shachter}, we have
\[ 
|\matW(i)| = \frac{1}{v_i}|\matW(i-1)| = \prod_{i=1}^n v_i^{-1} 
\] 
so that the determinant in \Eq{eq:p-ind-w1} factors as a function of 
$i$.  Also, \Eq{eq:shachter} implies (by induction) that each element 
$w_{ij}$ in $\matW$ is a sum of terms each being a function of $\vecb_i$ 
and $v_i$.  Consequently, the exponent in \Eq{eq:p-ind-w1} factors as a 
function of $i$.  Thus, given the Jacobian in \Eq{eq:jac-w}, which
also factors as a function of $i$, we obtain
\begin{equation} \label{eq:pinw2}
p(\vecv,\matB) = \prod_{i=1}^n p(v_i, \vecb_i)
\end{equation}
\Eqs{eq:pinw1} and \ref{eq:pinw2} imply the theorem. \qed

\section{From Consistency to Necessity} \label{sec:inev}

According to \Eq{eq:phidir3a}, when $p(\ThU|\hBsc)$ is Dirichlet, not
only are the parameters for each variable independent, but also the
parameters corresponding to each instance of every variable's parents
are independent.  We call this additional independence {\em local
parameter independence}, again after Spiegelhalter and Lauritzen
(1990).  From our discussion in the previous section, we see that the
Dirichlet assumption is consistent with likelihood equivalence and
both global and local parameter independence.

It is interesting to ask whether there are any other choices
for $p(\ThU|\hBsc)$ that are consistent in this sense.  In our
two-binary-variable example, using likelihood equivalence and
\Eq{eq:jac-d}, we obtain
\begin{equation} \label{eq:f1}
p(\ta{x}, \ta{y|x}, \ta{y|\bar{x}}|\hBxtoy) =
  \frac{\ta{x}(1-\ta{x})}{\ta{y}(1-\ta{y})} \ 
  p(\ta{y}, \ta{x|y}, \ta{x|\bar{y}}|\hBytox)
\end{equation}
where

\begin{equation} \label{eq:xyyx}
\ta{y}=\ta{x}\ta{y|x}+(1-\ta{x})\ta{y|\bar{x}}, \ \ \ 
\ta{x|y}=\frac{\ta{xy}}{\ta{x}\ta{y|x}+(1-\ta{x})\ta{y|\bar{x}}}, \ \ \ 
\ta{x|\bar{y}}=\frac{\ta{x\bar{y}}}{1-(\ta{x}\ta{y|x}+
  (1-\ta{x})\ta{y|\bar{x}})}
\end{equation}

\comment{
\begin{equation} \label{eq:xyyx}
\begin{array}{c}
\ta{y}=\ta{x}\ta{y|x}+(1-\ta{x})\ta{y|\bar{x}} \nonumber \\*[6pt]
\ta{x|y}=\frac{\ta{xy}}{\ta{x}\ta{y|x}+(1-\ta{x})\ta{y|\bar{x}}} 
  \\*[6pt]
\ta{x|\bar{y}}=\frac{\ta{x\bar{y}}}{1-(\ta{x}\ta{y|x}+
  (1-\ta{x})\ta{y|\bar{x}})} \nonumber
\end{array}
\end{equation}
}

\noindent
Applying global and local parameter independence to both sides of
\Eq{eq:f1}, we get
\begin{equation} \label{eq:f2}
f_x(\ta{x}) \ f_{y|x}(\ta{y|x}) \ f_{y|\bar{x}}(\ta{y|\bar{x}}) =
  \frac{\ta{x}(1-\ta{x})}{\ta{y}(1-\ta{y})} \ 
  f_y(\ta{y}) \ f_{x|y}(\ta{x|y}) \ f_{x|\bar{y}}(\ta{x|\bar{y}})
\end{equation}
where $f_x, f_{y|x}, f_{y|\bar{x}}, f_y, f_{x|y},$ and $f_{x|\bar{y}}$
are unknown pdfs. \Eqs{eq:xyyx} and \ref{eq:f2} define a {\em
functional equation}.  Geiger and Heckerman (1995a)\nocite{GH95tr-dir}
show that the only pdf solutions to \Eqs{eq:xyyx} and \ref{eq:f2} (and
their generalizations for non-binary variables) are those where
$p(\ThU|\hBsc)$ is a Dirichlet distribution.  Heckerman et al. (1995b)
generalize this result to $n$-variable models.  Thus, in the discrete
case, global and local parameter independence and likelihood
equivalence provide a characterization of the Dirichlet distribution.

Geiger and Heckerman (1995b)\nocite{GH95tr-nw} obtain an analogous
result for the two-variable linear-regression case.  In particular,
Let $\{m_1,v_1,m_{2|1},b_{12},v_{2|1}\}$ and
$\{m_2,v_2,m_{1|2},b_{21},v_{1|2}\}$ denote the parameters for the
structures $\X_1 \rightarrow \X_2$ and $\X_1
\leftarrow \X_2$, respectively.  As we demonstrated in
the previous section, if $p(\ThU|\hBsc)$ is a bivariate normal-Wishart
distribution, then global parameter independence holds.  Conversely,
assuming global parameter independence and likelihood equivalence, and
using the Jacobians in Equations~\ref{eq:jac-m} and
\ref{eq:jac-w}, we obtain the functional equation

\begin{equation} \label{eq:conj1}
f_1(m_1,v_1) \ f_{2|1}(m_{2|1}, b_{12},v_{2|1}) = \frac{v_1^2
  v_{2|1}^3}{v_2^2 v_{1|2}^3} f_2(m_2,v_2) \ f_{1|2}(m_{1|2},
  b_{21},v_{1|2})
\end{equation}

\comment{
\begin{eqnarray} \label{eq:conj1}
\nonumber
f_1(m_1,v_1) \ f_{2|1}(m_{2|1}, b_{12},v_{2|1}) = \frac{v_1^2
  v_{2|1}^3}{v_2^2 v_{1|2}^3} \cdot \\
\ f_2(m_2,v_2) \ f_{1|2}(m_{1|2},
  b_{21},v_{1|2})
\end{eqnarray}
}

\noindent
where $f_1$, $f_{2|1}$, $f_2$, and $f_{1|2}$ are arbitrary pdfs, and,
from \ref{eq:m-mu} and \Eqs{eq:shachter},
\[
v_2 = v_{2|1} + v_1 b_{12}^2 \ \ \ \ \ 
b_{21} = \frac{b_{12} v_1}{v_2} \ \ \ \ \ 
v_{1|2} = \frac{v_{2|1} v_1}{v_2}
\]
\[
m_2 = m_{2|1} + b_{12}m_1 \ \ \ \ \ 
m_{1|2} = m_1 + b_{21}m_2
\]
The only pdf solutions to this functional equation are those where
$p(\vecmu,\matW|\hBsc)$ is a bivariate normal-Wishart distribution
times an arbitrary pdf $f(w_{12})$, where $w_{12}$ is the off-diagonal
element of $\matW$.  Given the additional assumption of local
parameter independence, which in this context says that each set of
{\em standardized} parameters $\{m_1^*,v_1^*\}$, $\{m_{2|1}^*,
b_{12}^*, v_{2|1}^*\}$, $\{m_2^*,v_2^*)$, and $\{m_{1|2}^*, b_{21}^*,
v_{1|2}^*\}$ are mutually independent, the function $f(\cdot)$ must be
a constant.

In general, given likelihoods that satisfy covered-arc-reversal
equivalence, the assumptions of parameter independence in combination
with likelihood equivalence will yield a functional equation.  As in
these examples, the solutions to the equations may provide
characterizations of well-known distributions.  Alternatively, there
may be no solutions or new distribution classes may be revealed.

\section{Discussion}

Parameter independence, covered-arc-reversal equivalence, and
likelihood equivalence together yield strong constraints on priors.
In the discrete case, someone who adopts these assumptions can have
only one 
\textcolor{blue}{effective} sample size for all variables.  That is, this
person must be equally confident in his or her knowledge about each
variable.  Similarly, in the linear-regression case, these assumptions
permit the use of only two 
\textcolor{blue}{effective sample sizes: one for 
the normal component of the prior and one for the Wishart component.}
Nonetheless, if one learns about a portion of
domain by reading or through word of mouth, or simply by applying
common sense, then one or two 
\textcolor{blue}{effective} sample sizes will likely be
inadequate for an accurate expression of priors.

Thus, these assumptions should be checked when applying them to any
real-world problem.  If the assumptions are incorrect, then a
sensitivity analysis should be done to see if the violation of the
assumptions has much effect on the conclusions.  If there is a large
effect, then the assumptions should be relaxed.  A proposal for doing
so is given in Heckerman et al. (1995b).

\section*{Acknowledgments}

We thank Enrique Castillo, Clark Glymour, Chris Meek, Peter Spirtes,
Bo Thiesson, and anonymous reviewers for their useful suggestions.

\bibliographystyle{apalike}

\end{document}